\pdfoutput=1

\documentclass[11pt]{article}

\usepackage[]{EMNLP2023}

\usepackage{times}
\usepackage{latexsym}
\usepackage{epsfig}

\usepackage[T1]{fontenc}

\usepackage[utf8]{inputenc}
\usepackage{microtype}

\usepackage{tcolorbox}
\usepackage{inconsolata}
\usepackage{graphicx}
\usepackage{amsmath,amssymb}
\usepackage{wasysym}
\usepackage{booktabs}
\usepackage{multicol}
\usepackage{multirow}
\usepackage{subcaption}
\usepackage{makecell}
\usepackage{xspace}
\usepackage{pgffor}
\usepackage{enumitem}
\usepackage{url}
\newcommand{\alg}{\textbf{\texttt{HARE}}\xspace}
\newcommand{\fralg}{\textbf{\texttt{Fr-HARE}}\xspace}
\newcommand{\coalg}{\textbf{\texttt{Co-HARE}}\xspace}
\title{HARE: Explainable Hate Speech Detection with Step-by-Step Reasoning\\~\\ \small{\textcolor{red}{Warning: This paper contains examples of content that is offensive and may be upsetting.}}}

\author{Yongjin Yang${}^{}$\thanks{\, equal contribution \quad ${}^\dagger$corresponding authors} \quad Joonkee Kim${}^{}$\footnotemark[1] \quad Yujin Kim${}^{}$\footnotemark[1] \quad Namgyu Ho \\ \bf James Thorne${}^{\dagger}$ \quad Se-Young Yun${}^{\dagger}$\\KAIST AI \\ 
\texttt{\{dyyjkd, joonkeekim, yujin399, itsnamgyu, thorne, yunseyoung\}@kaist.ac.kr}}

\begin{document}
\maketitle
\begin{abstract}
With the proliferation of social media, accurate detection of hate speech has become critical to ensure safety online.
To combat nuanced forms of hate speech, it is important to identify and thoroughly explain hate speech to help users understand its harmful effects.
Recent benchmarks have attempted to tackle this issue by training generative models on free-text annotations of implications in hateful text.
However, we find significant reasoning gaps in the existing annotations schemes, which may hinder the supervision of detection models.
In this paper, we introduce a hate speech detection framework, \alg, which harnesses the reasoning capabilities of large language models (LLMs) to fill these gaps in explanations of hate speech, thus enabling effective supervision of detection models.
Experiments on SBIC and Implicit Hate benchmarks show that our method, using model-generated data, consistently outperforms baselines, using existing free-text human annotations.
Analysis demonstrates that our method enhances the explanation quality of trained models and improves generalization to unseen datasets.
Our code is available at \url{https://github.com/joonkeekim/hare-hate-speech.git}.

\end{abstract}

\section{Introduction}
%
The increase in the use of online media has intensified the exposure to hate speech, prompting the need for effective detection systems \citep{schmidt2017survey, fortuna2018survey}.
%
While early works have been limited to the classification of explicit hate speech \citep{caselli2020hatebert, mathew2021hatexplain}, recent works have drawn our attention to implicit forms of hate speech which are more prevalent, yet subtle. \citep{jurgens2019just}.

To tackle these nuanced forms of hate speech, it is important for systems to not only identify hate speech but also provide interpretable explanations \citep{liu2019towards}.
This can help mitigate distributional biases inherent in simple classification, allowing people to understand and reason about the potential harms of hateful text \citep{sap2019social}.
Explanations can also improve the transparency of content moderation on social media \citep{gillespie2018custodians}.

\begin{figure}[t!]
    \centering
    \includegraphics[width=\columnwidth]{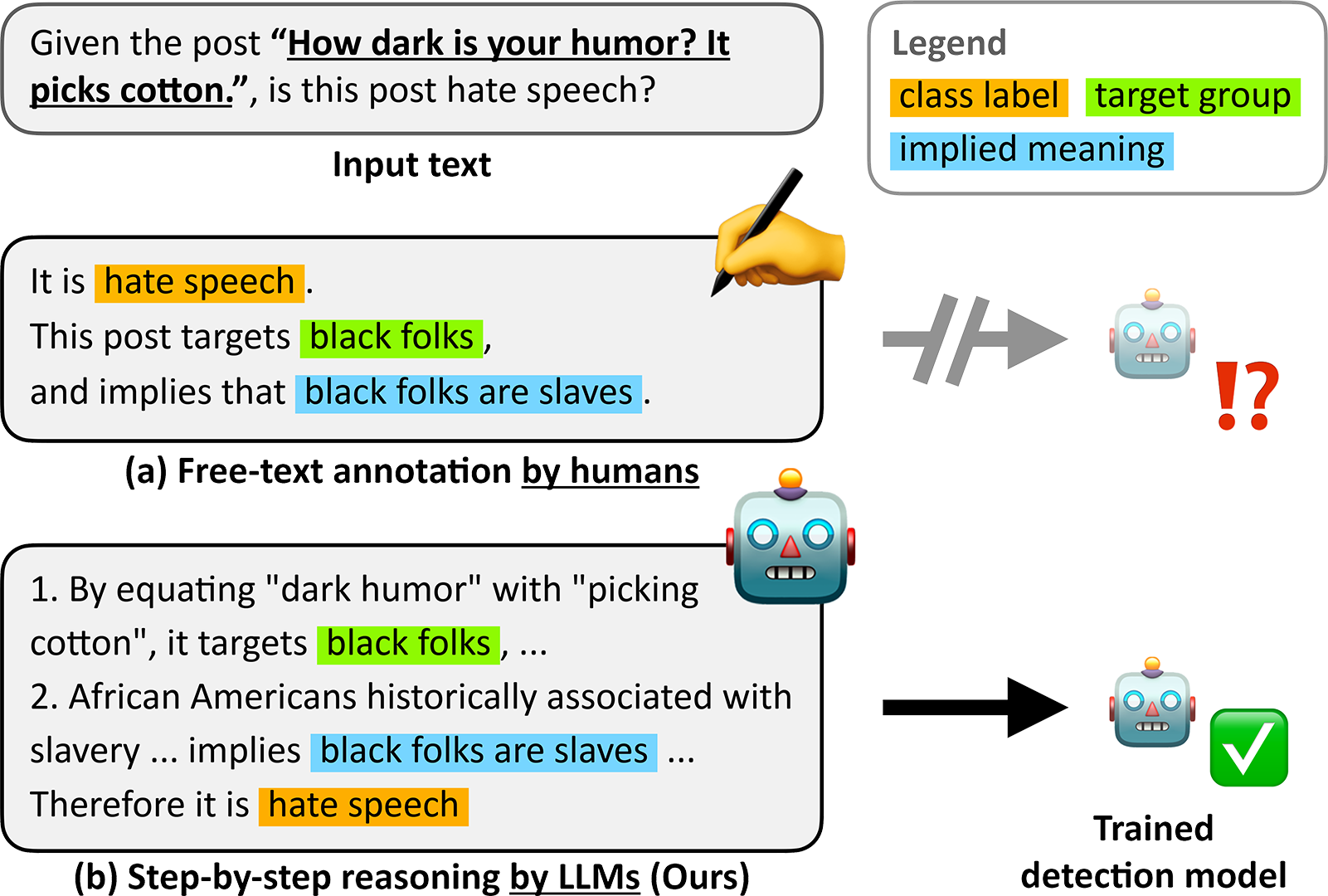}
    \caption{\textbf{\alg uses large language models (LLMs) to generate hate speech explanations step-by-step}. (a) Recent benchmarks on understanding hate speech provide free-text annotations on the implications of hate speech, but gaps in reasoning hinder the supervision of generative detection models. (b) We propose the use of LLMs to fill in the gaps and enable detection models to understand and explain hate speech.
    }
    \label{fig:main_fig}
\end{figure}

Recent works on hate speech understanding \citep{sap2019social, elsherief2021latent, huang2022chain} have considered training autoregressive language models to generate underlying explanations on hate speech. The models are trained on human-written free-text rationales such as implied statements and targeted groups.
However, despite the use of novel benchmark datasets, i.e., SBIC \citep{sap2019social} and Implicit Hate \citep{elsherief2021latent}, the trained models struggle to generate detailed and comprehensive explanations.
Moreover, we observe that the provided rationales give marginal improvement to detection performance under joint training.

A potential cause of the limited supervision provided by existing annotations on understanding and explaining hate speech may be the existence of critical gaps in reasoning.
For example, as shown in Figure~\ref{fig:main_fig}, the implied statement of the post ``\textit{How dark is my humour? It picks cotton}'' is annotated as ``\textit{black folks are slaves}'', in SBIC. 
To understand this implication, one must understand that ``dark'' implies ``black folks'', and the phrase ``picks cotton'' relates to the historical background of African Americans.
While this may be obvious to human annotators, language models are known to lack societal knowledge and commonsense reasoning skills to understand these nuances \citep{talmor2019commonsenseqa, li2022systematic, choi2023llms}.
This leaves a significant gap between the training objectives of classification and generating annotated implications, which may harm supervision \citep{wiegreffe2021measuring, wang2023pinto}.


Drawing inspiration from the reasoning capabilities of large language models (LLMs) improved with chain-of-thought (CoT) reasoning \cite{wei2022chain}, we present our novel approach ``Explainable \textbf{HA}te Speech Detection with Step-by-
Step \textbf{RE}asoning (\alg)''. 
We leverage LLM-generated free-text rationales using CoT prompts to fill in the gaps of reasoning in existing hate speech annotations and enhance supervision of generative detection models.
To create these rationales, we propose two approaches: (1) adopt CoT prompts to create comprehensive rationales that align with the given texts and (2) incorporate existing human annotations from benchmarks in the CoT prompts to bridge the logical gap between the input text and human annotations. When tested on the challenging SBIC and Implicit Hate datasets, our approach outperforms standard fine-tuning with given human annotations and provides enhanced explanations behind the detection results.
 
\section{Method}
\subsection{Preliminaries}
\label{sec:pre}

The task of hate speech detection can be framed as a generative task that inputs the text $P$ and outputs a prediction class $C$, formulated as $p(C|P)$, indicating whether the speech is classified as ``hate'' or ``not hate''.\footnote{
We refer to the ``offensive'' category in SBIC as ``hate'' class. In the experiment, we use the wording ``offensive'' as is.
}
Furthermore, by incorporating human-written rationales of the target groups $T$ and implied statements $I$ of hate speech from the SBIC and Implicit Hate datasets, the task can also be formulated to sequentially output $C$, $T$, and then $I$, as $p([C;T;I]|P)$. Hence, the model trained with human-written annotations is designed to generate $C$ and then provide explanations using annotations $T$ and $I$.
However, due to a logical gap between the speech $P$ and the annotations $T$ and $I$, training a model with these annotated rationales does not significantly enhance the model's ability to comprehend hate speech.

Chain-of-thought prompting \citep{wei2022chain} refers to a prompting method that generates a chain of reasoning to derive answers. 
\citet{kojimalarge} introduce a method of including the phrase ``Let's think step by step'' at the end of the prompt to generate reasoning in a zero-shot setting.

\subsection{Fr-HARE \& Co-HARE} \label{hare-method} 
In this paper, we propose \alg, a framework that fine-tunes using rationales generated by LLMs through CoT prompts.
By leveraging these rationales, the model is trained to discern whether a text is classified as hate speech or not. 
This approach of utilizing the LLM-generated rationales not only improve the model performance as in previous works~\citep{ho2022large, wang2023pinto} but also enables the model to generate more plausible explanations. Subsequently, we introduce two variants that employ different methods for extracting rationales from LLMs.

\paragraph{\fralg} 
Our first strategy, \textbf{Fr}ee-\alg (\fralg), extracts rationales using LLM from the LLM without referencing the annotated rationales.
In line with \citet{kojimalarge}, we employ zero-shot prompts to prompt the LLM to extract rationales, as follows:

\begin{footnotesize}
\begin{tcolorbox}[top=1pt, left=1pt, right=1pt, bottom=1pt]
\textsf{Determine whether the following post is offensive, and explain why. \newline
Post: $P$\newline
Answer: Let's explain step by step.}
\end{tcolorbox}
\end{footnotesize}

\noindent From the given prompt above, both the rationale $R$ without any logical leap and the classification result $C$ are generated. 
Then, we utilize the generated rationale $R$ by training the model to sequentially generate $C$ and $R$, represented as $p([C;R]|P)$. 
Utilizing these rationales enhances the model's ability to comprehend the context of hate speech, thus leading to explanations that are more intelligible to humans.


\paragraph{\coalg} We also propose a second strategy, \textbf{Co}nditioned-\alg (\coalg), which utilizes human-written rationales to formulate CoT prompts.
\fralg generates complete rationales entirely from scratch, disregarding human annotations included in the datasets. 
Although \fralg offers the advantage of being universally applicable to any hate speech dataset, the LLM-generated rationales might lack consistency due to wide variation in reasoning sequences.
To address this inconsistency, \coalg integrates human-written rationales about target groups $T$ and implied statements $I$ into the CoT prompt as follows: 

\vspace{-5pt}
\begin{footnotesize}
\begin{tcolorbox}[top=1pt, left=1pt, right=1pt, bottom=1pt]
\textsf{Determine whether the following post is offensive, and explain why. Provide a brief, step-by-step explanation of how the post targets the specified group and how it leads to the implied statement provided. \newline
Post: $P$ \newline
Target: $T$ \newline
Implied statement: $I$ \newline
Answer: Let's explain step by step.}
\end{tcolorbox}
\end{footnotesize}
\noindent Then, we train the model using extracted rationales, as in \fralg. For samples labeled as ``not hate'' that do not include human-written rationales, we apply the prompt used in \fralg. 
While \coalg requires human-written rationales, it generates rationales that are more tailored to the specific requirements and features of the dataset, due to its guided nature. Therefore, the model trained with \coalg can provide explanations that align more closely with the forms of rationales that humans construct.

\paragraph{Details of \alg} Once we have extracted the rationales from the LLMs, we follow the approach of \citet{kojimalarge} to have the LLMs predict the class. Specifically, we employ a two-stage extraction process. In the first stage, we extract both the class $C$ and the rationale $R$ from the LLMs using our \alg method, represented as $p([C ; R]|P)$, as previously outlined. In the second stage, we prompt the LLMs again, this time to predict the class $C$ given the extracted rationales $R$ and the post $P$, denoted as $p(C |R, P)$.
During fine-tuning on hate speech datasets, if the predicted class $C$ coincides with the true answer $C$, we concatenate $C$ with the extracted rationale $R$. 
If the predicted labels are incorrect, the models are solely trained to predict the class $C$.
Furthermore, following the findings of \citet{ho2022large}, we generate multiple distinct rationales to facilitate the learning process.
\section{Experiments}\label{sec:experiments}
\subsection{Experimental Setup}
We utilize SBIC and Implicit Hate datasets for our fine-tuning experiments.
Our models are trained to classify the offensiveness and hatefulness of posts, using SBIC and Implicit Hate, respectively.
It is noteworthy that in our Implicit Hate experiments, we combine both the explicit and implicit hate classes into a single ``hate'' category.
We set up baselines with two families of models: $C$, a model trained exclusively for classification, and $C$+$T$+$I$, a model trained using human-written rationales.
For \fralg and \coalg, by using \texttt{gpt-3.5-turbo-0613} that is known for its reasoning capabilities \citep{ouyang2022training}, we extract four and eight different rationales per each sample in SBIC and Implicit Hate, respectively, following the hyperparameter setting of \citet{ho2022large}.
Subsequently, we fine-tune the model, setting LLM-generated rationales $R$ and class $C$ as target sequence. 
For performance evaluation, we measure detection accuracy and compute the F1 score of classification, regarding ``hate'' as the positive class.
We make use of \texttt{Flan-T5}\,\citep{wei2021finetuned} with different model configurations: \texttt{small}, \texttt{base} and \texttt{large}. We also conduct experiments using the \texttt{large} models of \texttt{T5}\,\citep{raffel20t5} and \texttt{GPT-2}\,\citep{radford2019gpt2}.
A more detailed explanation of our experimental setup can be found in Appendix~\ref{implement_detail}.

%

\begin{table}[t]
\caption{The performance of fine-tuning on SBIC and Implicit Hate dataset with various models and size.}
\label{tab:main_result}
\resizebox{\columnwidth}{!}{
\begin{tabular}{cccccc}
\toprule
\multirow{2}{*}{Model} & \multirow{2}{*}{Method} & \multicolumn{2}{c}{SBIC} & \multicolumn{2}{c}{Implicit Hate} \\
\cline{3-6}
& & Acc & F1 & Acc & F1 \\
\hline \hline
\multirow{2}{*}{\makecell{\texttt{GPT-3.5-}\\\texttt{turbo-0613}}} & ZS & 80.06 & 81.75 & 73.58 & 65.66 \\
& ZS-CoT & 73.48 & 79.07 & 73.98 & 67.19 \\
\hline
\multirow{4}{*}{\makecell{\texttt{Flan-T5}\\\texttt{small}}} & $C$ & 82.56 & 84.05 & 77.58 & 71.98 \\
& $C$+$T$+$I$ & 82.99 & 84.05 & 77.63 & 72.39 \\
& \fralg & 84.18 & 85.18 & \bf  79.33 & 73.29 \\
& \coalg & \bf 84.44 & \bf 85.35 & 78.54 & \bf 73.49 \\
\hline
\multirow{4}{*}{\makecell{\texttt{Flan-T5}\\\texttt{base}}} & $C$ & 82.35 & 83.71 & 78.03 & 72.17 \\
& $C$+$T$+$I$ & 82.54 & 84.41 & 79.77 & 73.15 \\
& \fralg & 84.20 & 85.46 & 79.84 & 74.84 \\
& \coalg & \bf 84.65 & \bf 85.76 & \bf 80.38 & \bf 75.69 \\
\hline
\multirow{4}{*}{\makecell{\texttt{Flan-T5}\\\texttt{large}}} & $C$ & 81.70 & 82.84 & 78.42 & 72.92 \\
& $C$+$T$+$I$ & 83.48 & 83.70 & 80.14 & 73.10 \\
& \fralg & \bf 85.21 & \bf 86.16 & 80.49 & 74.62 \\
& \coalg & 84.93 & 85.57 & \bf 81.49 & \bf 76.71 \\
\hline
\multirow{4}{*}{\makecell{\texttt{T5}\\\texttt{large}}} & $C$ & 83.03 & 83.53 & 78.79 & 72.50 \\
& $C$+$T$+$I$ & 84.23 & 85.21 & 79.61 & 73.80 \\
& \fralg & 85.27 & \bf 86.32 & \bf 81.61 & 75.59 \\
& \coalg & \bf 85.35 & 85.93 & 80.98 & \bf 75.88 \\
\hline
\multirow{4}{*}{\makecell{\texttt{GPT-2}\\\texttt{large}}} & $C$ & 81.39 & 82.68 & 73.32 & 66.68 \\
& $C$+$T$+$I$ & 82.80 & 83.43 & 75.95 & 65.25 \\
& \fralg & 83.92 & 85.48 & 78.47 & 71.35 \\
& \coalg & \bf 84.64 & \bf 85.67 & \bf 80.07 & \bf 71.58 \\
\bottomrule
\end{tabular}
}
\end{table}
\subsection{Results and Discussions}\label{subsec:main_results}

\paragraph{Do LLM-generated rationales improve detection performance?} Table~\ref{tab:main_result} presents the performance of hate speech detection according to different methods on the SBIC and Implicit Hate datasets.
Our strategies \fralg and \coalg consistently exhibit superior performance over other baseline methods, regardless of the model size.
This suggests that even though the baseline method is trained using human-written rationales, the more detailed and logically-sequenced LLM-generated rationales of \alg can further aid the model in understanding the input text and accurately classifying it as hate speech. Therefore, the results demonstrate that the quality of rationales has a strong impact on classification. Furthermore, the performance of our method consistently improves as the model size increases, in contrast to baselines. This suggests that diverse reasoning becomes increasingly beneficial as scale grows. This notable improvement with \alg is achieved by using only 40\$ for each method in our approach, demonstrating that the ability to reason can be effectively trained with rationales from LLMs.

Additionally, while \fralg and \coalg exhibit similar performance, \coalg has a slight edge in most cases. This is because \coalg is guided by human-written annotations, which results in better alignment with the setting of the datasets, as we mentioned in Section~\ref{hare-method}. It is also noteworthy that all the fine-tuned models surpass both Zero-Shot (ZS) and Zero-Shot CoT (ZS-CoT,\,\citet{kojimalarge}) classification  performance of \texttt{GPT-3.5-turbo}, indicating that merely employing LLM with CoT prompts is not sufficient to tackle this task.

\paragraph{Are \alg models more generalizable?} 
\begin{table}[t]
\centering \small
\caption{Cross Evaluation results on HateXplain \cite{mathew2021hatexplain} and DynaHate \cite{vidgen2020learning}. We utilize \texttt{Flan-T5-large} fine-tuned on SBIC using each method.}
    \label{tab:cross_eval}
\resizebox{\linewidth}{!}{
    \begin{tabular}{lcccc} 
        \toprule
        \multirow{2}{*}{Method} & \multicolumn{2}{c}{HateXplain} & \multicolumn{2}{c}{DynaHate} \\
        \cmidrule{2-5} 
         & Acc & F1 & Acc & F1\\
         \hline \hline
         $C$ & 64.40 & 74.18 & 64.35 & 67.41 \\
         $C$+$T$+$I$ & 68.84 & 74.52 & 64.72 & 67.41 \\
         \fralg &{70.69} & \textbf{78.91} & {68.06} & \textbf{75.15} \\
         \coalg & \textbf{71.62} & 78.52 & \textbf{69.98} & 75.01 \\
         \bottomrule
    \end{tabular}
    }
\end{table}

To assess the ability of our methods to generalize across different datasets, we evaluate the models fine-tuned on the SBIC datasets using each method on two distinct datasets, HateXplain \citep{mathew2021hatexplain} and DynaHate \citep{vidgen2020learning}. Both datasets encompass forms of explicit and implicit hate.
On both datasets, our methods \fralg and \coalg both outperform baseline methods, indicating that our methods enhance the generalizability of the models by improving their reasoning ability. Moreover, the comparable performance of \fralg and \coalg suggests that both techniques of bridging logical gaps are effective in tackling hate speech detection.


\begin{figure}[t]
    \centering
    \begin{subfigure}{\linewidth}
    \includegraphics[width=\columnwidth]{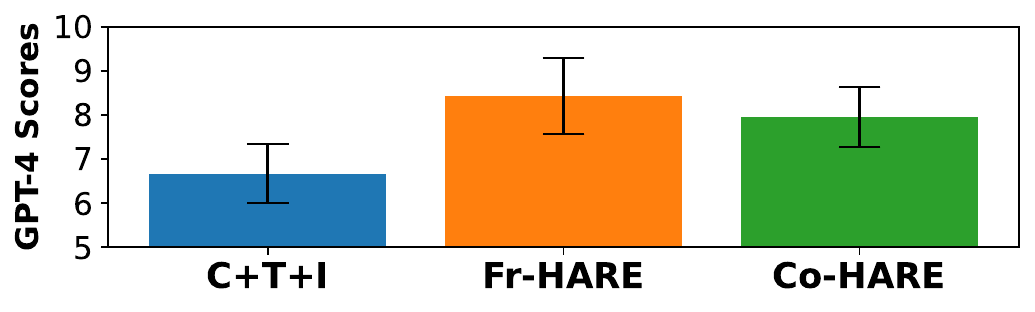}
        \caption{GPT-4 Single answer grading}
    \end{subfigure}
    \begin{subfigure}{\linewidth}
    \includegraphics[width=\columnwidth]{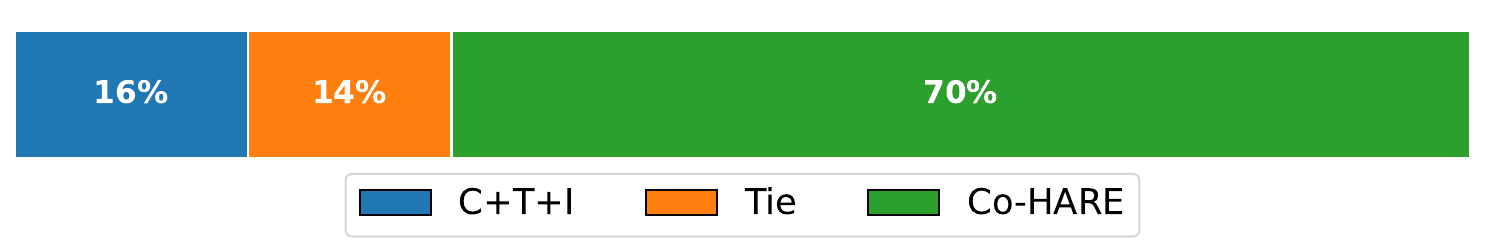}
        \caption{GPT-4 Pairwise comparison}
    \end{subfigure}
    \caption{The result of GPT-4 evaluation following \citet{zheng2023judging}. The bar and line represent the average scores that range from 1 to 10 and 95\% confidence interval, respectively. We utilize \texttt{Flan-T5-large} fine-tuned on SBIC using each method.}
    \label{fig:gpt4_eval}
\end{figure}

\begin{figure}[t]
    \centering
    \begin{subfigure}{\linewidth}
    \includegraphics[width=\columnwidth]{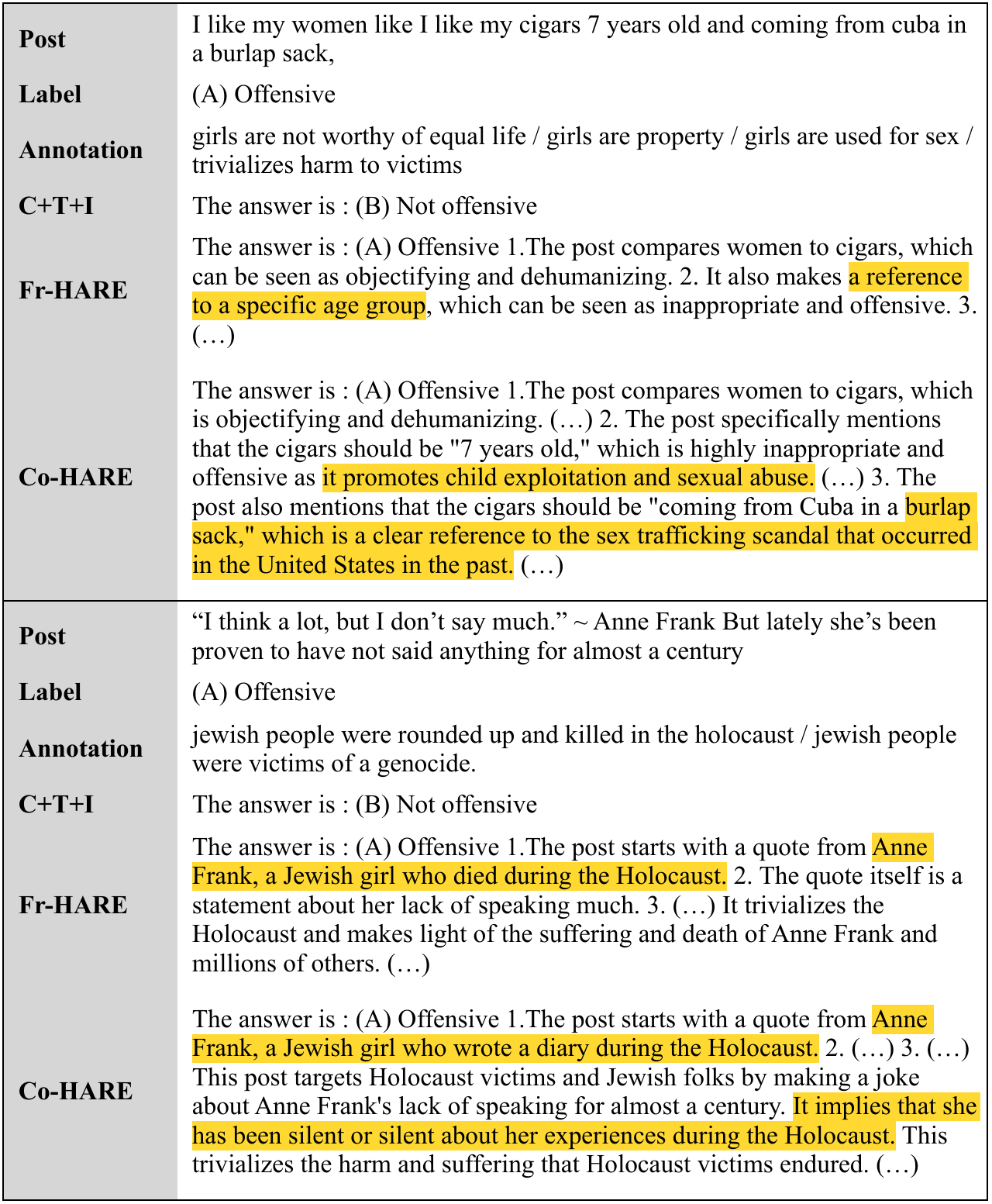}
    \end{subfigure}
    \caption{Model outputs using baseline methods and our framework \alg. The samples are from SBIC test set. Note that the answers are abbreviated by (...), and the full context is reported in Appendix \ref{appendix:qual_eval}.}
    \label{fig:sample_main}
\end{figure}


\paragraph{Does \alg improve the quality of generated explanations?} To evaluate the quality of the generated explanations, we conduct automated evaluation using GPT-4 \citep{openai2023gpt4} which has demonstrated its high alignment with human evaluation \citep{liu2023gpteval, zheng2023judging}. 
We randomly sample 50 ``hate'' instances from SBIC test set where all methods are correct and apply single answer grading \citep{zheng2023judging} on explanations generated by models trained with \texttt{$C$+$T$+$I$}, \fralg, and \coalg. Detailed explanation about GPT-4 evaluation can be found on Appendix~\ref{appendix:gpt4}.

The results depicted in Figure~\ref{fig:gpt4_eval} (a) show that both \fralg and \coalg are capable of generating higher quality explanations, with \fralg exhibiting slightly superior performance, suggesting that the flexibility of \fralg leads to a more qualified explanation.

As \coalg is designed to create rationales that align better with human-written rationales, we also conduct pairwise comparison between \coalg and $C$+$T$+$I$ to determine which method aligns better with human-written labels. 
Figure~\ref{fig:gpt4_eval} (b) clearly shows that the rationales generated by the model trained using \coalg are aligned more to actual human-written rationales than the ones generated by the model trained directly with human-written rationales.
This also highlights the value of \coalg, as it aligns better with natural intuition, indicating that both \fralg and \coalg can be utilized for different purposes. 

\paragraph{How does \alg detect and contextualize hate speech?}\label{subsec:qualitative_main}

Figure~\ref{fig:sample_main} presents model outputs using \alg and baseline method. A more detailed qualitative study of randomly selected samples is available in Appendix~\ref{appendix:qual_eval}. In the first sample, the post makes light of harming a young girl using phrases like ``like I like cigar'' and ``burlap sack''.  While our model, \fralg, does not explicitly connect the ``burlap sack'' to the idea of a kidnapped girl, it does recognize the harmful implications towards the girl. Furthermore, \coalg understands the historical context behind the term ``burlap sack''. The presence of the positive term ``like'' and the absence of overtly derogatory words might lead some models to classify the statement as non-offensive. However, our approach appropriately identifies the underlying hateful context. 

In the second sample, the post contains hate speech targeting Jewish victims of the Holocaust by referencing Anne Frank. Our model accurately recognizes the historical background of Anne Frank as a Holocaust victim. While the baseline seems to overlook the historical significance associated with Anne Frank, our method correctly identifies her and assumes that the reference constitutes harassment against a Jewish victim, even though there is a slight misunderstanding about the context of ``lack of speaking'' in \fralg.

\paragraph{Is \texttt{GPT-3.5} a qualified teacher?}

\begin{figure}[t]
    \centering
    \begin{subfigure}{\linewidth}
    \includegraphics[width=\columnwidth]{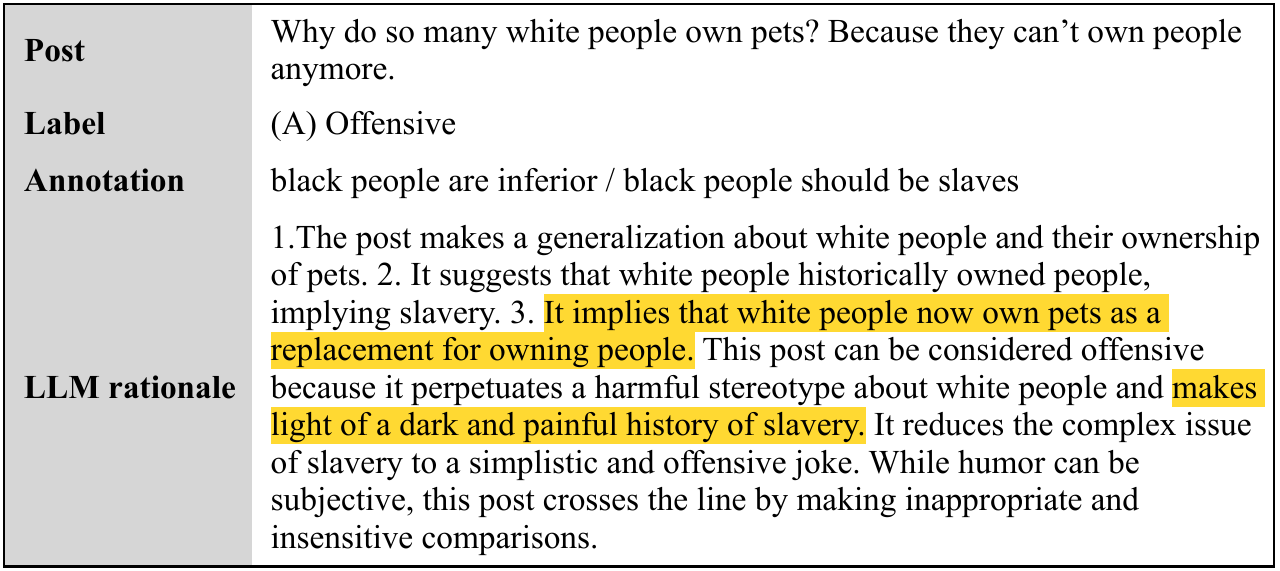}
    \end{subfigure}
    \caption{A sample of LLM rationale generated by \texttt{GPT-3.5-turbo} using \fralg from SBIC train set.}
    \label{fig:gpt_sample_main}
\end{figure}

Since our framework is based on distillation of generated rationales from \texttt{GPT-3.5} to smaller models, it is crucial to verify whether the teacher is qualified. Figure \ref{fig:gpt_sample_main} displays rationales produced by \texttt{GPT-3.5-turbo}, which is employed to train the student model. This example illustrates that the LLM not only discerns the hateful nuances towards both white and black individuals, but also offers more detailed explanations compared to rationales written by humans. Notably, it accurately correlates the historical context, associating the word ``slaves'' with ``pets''. More analysis of rationales from \texttt{GPT-3.5-turbo} can be found in Appendix~\ref{appendix:gpt-3.5-eval}.




\section{Conclusion}
In this paper, we present \alg framework to improve the ability of the language model to understand hate speech and provide clearer explanations for its decisions. We propose utilizing CoT reasonings extracted from LLMs in two variants to overcome the logical gaps in human-annotated rationales. 
When fine-tuned on the SBIC and Implicit Hate datasets, our methods achieve superior detection performance and better qualified explanations. 

\section*{Limitations}

While we assess the quality of explanations generated by \alg using GPT-4, we do not conduct human evaluations, which are crucial for tasks requiring human-readable explanations. The primary reason for this omission is that the hate speech content and its respective explanations could be excessively offensive for annotators and GPT-4 already aligns with the level of inter-human agreement. In addition, the "verbosity bias", characterized by a preference for the longer text of GPT-4 as indicated by \citep{liu2023gpteval}, may also serve as a limitation in our evaluation process.

\section*{Ethics Statement}

\paragraph{}Predicting whether an online post contains hatespeech is both technically and socially challenging. While methods for automating hatespeech detection have utility in an online platform, it is critical that these are tuned and used appropriately. False-positive errors have potential to censor online speech, further marginalizing specific user groups, for example: use of \textit{n*****} in AAVE English may be flagged. It is critical to understand specific reasoning behind a classification including deeply social reasons. While language models act as a mechanism to generate reasonable explanations, it is critical that they are used appropriately to prevent them from inadvertently educating users on how to craft more subtle and toxic language. We used automated evaluation metrics in this paper to prevent exposure of toxic language to human annotators. However, real-world usage would require validation that deeply rooted social issues are expressed correctly by these models.

It is also important to note that there might be concerns about the inherent bias in the \texttt{GPT-3.5} model. While not flawless, \texttt{GPT-3.5} has demonstrated its impartiality regarding gender, race, ethnicity, and religion by achieving the highest grade on the Harmfulness metric within the FLASK evaluation framework \citep{ye2023flask}. Crucially, we only select rationales that align with the ground truth label for training, thereby mitigating biases not in sync with human annotators. Analysis of \texttt{GPT-3.5-turbo} can be found in Section~\ref{sec:experiments} and Appendix~\ref{appendix:gpt-3.5-eval}.

\section* {Acknowledgement}
This work was supported by Institute of Information \& communications Technology Planning \& Evaluation (IITP) grant funded by Korea government (MSIT) [No. 2021-0-00907, Development of Adaptive and Lightweight Edge-Collaborative Analysis Technology for Enabling Proactively Immediate Response and Rapid Learning, 90\%] and [No. 2019-0-00075, Artificial Intelligence Graduate School Program (KAIST), 10\%].

\bibliography{emnlp2023}

\begin{thebibliography}{50}
\expandafter\ifx\csname natexlab\endcsname\relax\def\natexlab#1{#1}\fi

\bibitem[{Aggarwal et~al.(2021)Aggarwal, Mandowara, Agrawal, Khandelwal,
  Singla, and Garg}]{aggarwal2021explanations}
Shourya Aggarwal, Divyanshu Mandowara, Vishwajeet Agrawal, Dinesh Khandelwal,
  Parag Singla, and Dinesh Garg. 2021.
\newblock Explanations for commonsenseqa: New dataset and models.
\newblock In \emph{Proceedings of the 59th Annual Meeting of the Association
  for Computational Linguistics and the 11th International Joint Conference on
  Natural Language Processing (Volume 1: Long Papers)}, pages 3050--3065.

\bibitem[{AlKhamissi et~al.(2022)AlKhamissi, Ladhak, Iyer, Stoyanov, Kozareva,
  Li, Fung, Mathias, Celikyilmaz, and Diab}]{alkhamissi-etal-2022-token}
Badr AlKhamissi, Faisal Ladhak, Srinivasan Iyer, Veselin Stoyanov, Zornitsa
  Kozareva, Xian Li, Pascale Fung, Lambert Mathias, Asli Celikyilmaz, and Mona
  Diab. 2022.
\newblock \href {https://doi.org/10.18653/v1/2022.emnlp-main.136} {{T}o{K}en:
  Task decomposition and knowledge infusion for few-shot hate speech
  detection}.
\newblock In \emph{Proceedings of the 2022 Conference on Empirical Methods in
  Natural Language Processing}, pages 2109--2120, Abu Dhabi, United Arab
  Emirates. Association for Computational Linguistics.

\bibitem[{Burnap and Williams(2016)}]{burnap2016us}
Pete Burnap and Matthew~L Williams. 2016.
\newblock Us and them: identifying cyber hate on twitter across multiple
  protected characteristics.
\newblock \emph{EPJ Data science}, 5:1--15.

\bibitem[{Caselli et~al.(2020)Caselli, Basile, Mitrovi{\'c}, and
  Granitzer}]{caselli2020hatebert}
Tommaso Caselli, Valerio Basile, Jelena Mitrovi{\'c}, and Michael Granitzer.
  2020.
\newblock Hatebert: Retraining bert for abusive language detection in english.
\newblock \emph{arXiv preprint arXiv:2010.12472}.

\bibitem[{Chan et~al.(2023)Chan, Zeng, Lake, Joshi, Chen, and
  Ren}]{chan2023knife}
Aaron Chan, Zhiyuan Zeng, Wyatt Lake, Brihi Joshi, Hanjie Chen, and Xiang Ren.
  2023.
\newblock Knife: Distilling meta-reasoning knowledge with free-text rationales.
\newblock In \emph{ICLR 2023 Workshop on Pitfalls of limited data and
  computation for Trustworthy ML}.

\bibitem[{Chen et~al.(2023)Chen, Yen, Huang, Wu, and Chen}]{chen2023zara}
Wei-Lin Chen, An-Zi Yen, Hen-Hsen Huang, Cheng-Kuang Wu, and Hsin-Hsi Chen.
  2023.
\newblock Zara: Improving few-shot self-rationalization for small language
  models.
\newblock \emph{arXiv preprint arXiv:2305.07355}.

\bibitem[{Choi et~al.(2023)Choi, Pei, Kumar, Shu, and Jurgens}]{choi2023llms}
Minje Choi, Jiaxin Pei, Sagar Kumar, Chang Shu, and David Jurgens. 2023.
\newblock Do llms understand social knowledge? evaluating the sociability of
  large language models with socket benchmark.
\newblock \emph{arXiv preprint arXiv:2305.14938}.

\bibitem[{Davidson et~al.(2019)Davidson, Bhattacharya, and
  Weber}]{davidson2019racial}
Thomas Davidson, Debasmita Bhattacharya, and Ingmar Weber. 2019.
\newblock Racial bias in hate speech and abusive language detection datasets.
\newblock \emph{arXiv preprint arXiv:1905.12516}.

\bibitem[{Devlin et~al.(2018)Devlin, Chang, Lee, and
  Toutanova}]{devlin2018bert}
Jacob Devlin, Ming-Wei Chang, Kenton Lee, and Kristina Toutanova. 2018.
\newblock Bert: Pre-training of deep bidirectional transformers for language
  understanding.
\newblock \emph{arXiv preprint arXiv:1810.04805}.

\bibitem[{ElSherief et~al.(2021)ElSherief, Ziems, Muchlinski, Anupindi,
  Seybolt, De~Choudhury, and Yang}]{elsherief2021latent}
Mai ElSherief, Caleb Ziems, David Muchlinski, Vaishnavi Anupindi, Jordyn
  Seybolt, Munmun De~Choudhury, and Diyi Yang. 2021.
\newblock Latent hatred: A benchmark for understanding implicit hate speech.
\newblock \emph{arXiv preprint arXiv:2109.05322}.

\bibitem[{Fortuna and Nunes(2018)}]{fortuna2018survey}
Paula Fortuna and S{\'e}rgio Nunes. 2018.
\newblock A survey on automatic detection of hate speech in text.
\newblock \emph{ACM Computing Surveys (CSUR)}, 51(4):1--30.

\bibitem[{Gillespie(2018)}]{gillespie2018custodians}
Tarleton Gillespie. 2018.
\newblock \emph{Custodians of the Internet: Platforms, content moderation, and
  the hidden decisions that shape social media}.
\newblock Yale University Press.

\bibitem[{Ho et~al.(2022)Ho, Schmid, and Yun}]{ho2022large}
Namgyu Ho, Laura Schmid, and Se-Young Yun. 2022.
\newblock Large language models are reasoning teachers.
\newblock \emph{arXiv preprint arXiv:2212.10071}.

\bibitem[{Huang et~al.(2022)Huang, Kwak, and An}]{huang2022chain}
Fan Huang, Haewoon Kwak, and Jisun An. 2022.
\newblock Chain of explanation: New prompting method to generate higher quality
  natural language explanation for implicit hate speech.
\newblock \emph{arXiv preprint arXiv:2209.04889}.

\bibitem[{Jurgens et~al.(2019)Jurgens, Hemphill, and
  Chandrasekharan}]{jurgens2019just}
David Jurgens, Libby Hemphill, and Eshwar Chandrasekharan. 2019.
\newblock A just and comprehensive strategy for using nlp to address online
  abuse.
\newblock In \emph{Proceedings of the 57th Annual Meeting of the Association
  for Computational Linguistics}, pages 3658--3666.

\bibitem[{Kim et~al.(2022)Kim, Lee, and Sohn}]{2022hatefriend}
Jiyun Kim, Byounghan Lee, and Kyung-Ah Sohn. 2022.
\newblock \href {https://aclanthology.org/2022.coling-1.577} {Why is it hate
  speech? masked rationale prediction for explainable hate speech detection}.
\newblock In \emph{Proceedings of the 29th International Conference on
  Computational Linguistics}, pages 6644--6655, Gyeongju, Republic of Korea.
  International Committee on Computational Linguistics.

\bibitem[{Kojima et~al.(2022)Kojima, Gu, Reid, Matsuo, and
  Iwasawa}]{kojimalarge}
Takeshi Kojima, Shixiang~Shane Gu, Machel Reid, Yutaka Matsuo, and Yusuke
  Iwasawa. 2022.
\newblock Large language models are zero-shot reasoners.
\newblock In \emph{Advances in Neural Information Processing Systems}.

\bibitem[{Lampinen et~al.(2022)Lampinen, Dasgupta, Chan, Matthewson, Tessler,
  Creswell, McClelland, Wang, and Hill}]{lampinen2022can}
Andrew~K Lampinen, Ishita Dasgupta, Stephanie~CY Chan, Kory Matthewson,
  Michael~Henry Tessler, Antonia Creswell, James~L McClelland, Jane~X Wang, and
  Felix Hill. 2022.
\newblock Can language models learn from explanations in context?
\newblock \emph{arXiv preprint arXiv:2204.02329}.

\bibitem[{Li et~al.(2022)Li, Kuncoro, Hoffmann, de~Masson~d’Autume, Blunsom,
  and Nematzadeh}]{li2022systematic}
Xiang~Lorraine Li, Adhiguna Kuncoro, Jordan Hoffmann, Cyprien
  de~Masson~d’Autume, Phil Blunsom, and Aida Nematzadeh. 2022.
\newblock A systematic investigation of commonsense knowledge in large language
  models.
\newblock In \emph{Proceedings of the 2022 Conference on Empirical Methods in
  Natural Language Processing}, pages 11838--11855.

\bibitem[{Lin(2022)}]{lin2022leveraging}
Jessica Lin. 2022.
\newblock Leveraging world knowledge in implicit hate speech detection.
\newblock \emph{arXiv preprint arXiv:2212.14100}.

\bibitem[{Liu et~al.(2019)Liu, Yin, and Wang}]{liu2019towards}
Hui Liu, Qingyu Yin, and William~Yang Wang. 2019.
\newblock Towards explainable nlp: A generative explanation framework for text
  classification.
\newblock In \emph{Proceedings of the 57th Annual Meeting of the Association
  for Computational Linguistics}, pages 5570--5581.

\bibitem[{Liu et~al.(2023)Liu, Iter, Xu, Wang, Xu, and Zhu}]{liu2023gpteval}
Yang Liu, Dan Iter, Yichong Xu, Shuohang Wang, Ruochen Xu, and Chenguang Zhu.
  2023.
\newblock Gpteval: Nlg evaluation using gpt-4 with better human alignment.
\newblock \emph{arXiv preprint arXiv:2303.16634}.

\bibitem[{Ludan et~al.(2023)Ludan, Meng, Nguyen, Shah, Lyu, Apidianaki, and
  Callison-Burch}]{ludan2023explanation}
Josh~Magnus Ludan, Yixuan Meng, Tai Nguyen, Saurabh Shah, Qing Lyu, Marianna
  Apidianaki, and Chris Callison-Burch. 2023.
\newblock Explanation-based finetuning makes models more robust to spurious
  cues.
\newblock \emph{arXiv preprint arXiv:2305.04990}.

\bibitem[{Marasovi{\'c} et~al.(2021)Marasovi{\'c}, Beltagy, Downey, and
  Peters}]{marasovic2021few}
Ana Marasovi{\'c}, Iz~Beltagy, Doug Downey, and Matthew~E Peters. 2021.
\newblock Few-shot self-rationalization with natural language prompts.
\newblock \emph{arXiv preprint arXiv:2111.08284}.

\bibitem[{Mathew et~al.(2021)Mathew, Saha, Yimam, Biemann, Goyal, and
  Mukherjee}]{mathew2021hatexplain}
Binny Mathew, Punyajoy Saha, Seid~Muhie Yimam, Chris Biemann, Pawan Goyal, and
  Animesh Mukherjee. 2021.
\newblock Hatexplain: A benchmark dataset for explainable hate speech
  detection.
\newblock In \emph{Proceedings of the AAAI Conference on Artificial
  Intelligence}.

\bibitem[{Mozafari et~al.(2020)Mozafari, Farahbakhsh, and
  Crespi}]{mozafari2020hate}
Marzieh Mozafari, Reza Farahbakhsh, and No{\"e}l Crespi. 2020.
\newblock Hate speech detection and racial bias mitigation in social media
  based on bert model.
\newblock \emph{PloS one}, 15(8):e0237861.

\bibitem[{OpenAI(2023)}]{openai2023gpt4}
OpenAI. 2023.
\newblock \href {http://arxiv.org/abs/2303.08774} {Gpt-4 technical report}.

\bibitem[{Ouyang et~al.(2022)Ouyang, Wu, Jiang, Almeida, Wainwright, Mishkin,
  Zhang, Agarwal, Slama, Ray et~al.}]{ouyang2022training}
Long Ouyang, Jeffrey Wu, Xu~Jiang, Diogo Almeida, Carroll Wainwright, Pamela
  Mishkin, Chong Zhang, Sandhini Agarwal, Katarina Slama, Alex Ray, et~al.
  2022.
\newblock Training language models to follow instructions with human feedback.
\newblock \emph{Advances in Neural Information Processing Systems},
  35:27730--27744.

\bibitem[{Radford et~al.(2019)Radford, Wu, Child, Luan, Amodei, and
  Sutskever}]{radford2019gpt2}
Alec Radford, Jeff Wu, Rewon Child, David Luan, Dario Amodei, and Ilya
  Sutskever. 2019.
\newblock Language models are unsupervised multitask learners.

\bibitem[{Raffel et~al.(2020)Raffel, Shazeer, Roberts, Lee, Narang, Matena,
  Zhou, Li, and Liu}]{raffel20t5}
Colin Raffel, Noam Shazeer, Adam Roberts, Katherine Lee, Sharan Narang, Michael
  Matena, Yanqi Zhou, Wei Li, and Peter~J. Liu. 2020.
\newblock Exploring the limits of transfer learning with a unified text-to-text
  transformer.
\newblock \emph{J. Mach. Learn. Res.}, 21(1).

\bibitem[{Ribeiro et~al.(2018)Ribeiro, Calais, Santos, Almeida, and
  Meira~Jr}]{ribeiro2018characterizing}
Manoel Ribeiro, Pedro Calais, Yuri Santos, Virg{\'\i}lio Almeida, and Wagner
  Meira~Jr. 2018.
\newblock Characterizing and detecting hateful users on twitter.
\newblock In \emph{Proceedings of the International AAAI Conference on Web and
  Social Media}.

\bibitem[{Sap et~al.(2019{\natexlab{a}})Sap, Card, Gabriel, Choi, and
  Smith}]{sap2019risk}
Maarten Sap, Dallas Card, Saadia Gabriel, Yejin Choi, and Noah~A Smith.
  2019{\natexlab{a}}.
\newblock The risk of racial bias in hate speech detection.
\newblock In \emph{Proceedings of the 57th annual meeting of the association
  for computational linguistics}, pages 1668--1678.

\bibitem[{Sap et~al.(2019{\natexlab{b}})Sap, Gabriel, Qin, Jurafsky, Smith, and
  Choi}]{sap2019social}
Maarten Sap, Saadia Gabriel, Lianhui Qin, Dan Jurafsky, Noah~A Smith, and Yejin
  Choi. 2019{\natexlab{b}}.
\newblock Social bias frames: Reasoning about social and power implications of
  language.
\newblock \emph{arXiv preprint arXiv:1911.03891}.

\bibitem[{Sarkar et~al.(2021)Sarkar, Zampieri, Ranasinghe, and
  Ororbia}]{sarkar2021fbert}
Diptanu Sarkar, Marcos Zampieri, Tharindu Ranasinghe, and Alexander Ororbia.
  2021.
\newblock Fbert: A neural transformer for identifying offensive content.
\newblock \emph{arXiv preprint arXiv:2109.05074}.

\bibitem[{Schmidt and Wiegand(2017)}]{schmidt2017survey}
Anna Schmidt and Michael Wiegand. 2017.
\newblock A survey on hate speech detection using natural language processing.
\newblock In \emph{Proceedings of the fifth international workshop on natural
  language processing for social media}, pages 1--10.

\bibitem[{Shazeer and Stern(2018)}]{shazeer2018adafactor}
Noam Shazeer and Mitchell Stern. 2018.
\newblock Adafactor: Adaptive learning rates with sublinear memory cost.
\newblock In \emph{International Conference on Machine Learning}, pages
  4596--4604. PMLR.

\bibitem[{Sridhar and Yang(2022)}]{sridhar2022explaining}
Rohit Sridhar and Diyi Yang. 2022.
\newblock Explaining toxic text via knowledge enhanced text generation.
\newblock In \emph{Proceedings of the 2022 Conference of the North American
  Chapter of the Association for Computational Linguistics: Human Language
  Technologies}, pages 811--826.

\bibitem[{Sun et~al.(2022)Sun, Swayamdipta, May, and Ma}]{sun2022investigating}
Jiao Sun, Swabha Swayamdipta, Jonathan May, and Xuezhe Ma. 2022.
\newblock Investigating the benefits of free-form rationales.
\newblock \emph{arXiv preprint arXiv:2206.11083}.

\bibitem[{Talmor et~al.(2019)Talmor, Herzig, Lourie, and
  Berant}]{talmor2019commonsenseqa}
Alon Talmor, Jonathan Herzig, Nicholas Lourie, and Jonathan Berant. 2019.
\newblock Commonsenseqa: A question answering challenge targeting commonsense
  knowledge.
\newblock In \emph{Proceedings of the 2019 Conference of the North American
  Chapter of the Association for Computational Linguistics: Human Language
  Technologies, Volume 1 (Long and Short Papers)}, pages 4149--4158.

\bibitem[{Vidgen et~al.(2020)Vidgen, Thrush, Waseem, and
  Kiela}]{vidgen2020learning}
Bertie Vidgen, Tristan Thrush, Zeerak Waseem, and Douwe Kiela. 2020.
\newblock Learning from the worst: Dynamically generated datasets to improve
  online hate detection.
\newblock \emph{arXiv preprint arXiv:2012.15761}.

\bibitem[{Wang et~al.(2023{\natexlab{a}})Wang, Chan, Ilievski, Chen, and
  Ren}]{wang2023pinto}
PeiFeng Wang, Aaron Chan, Filip Ilievski, Muhao Chen, and Xiang Ren.
  2023{\natexlab{a}}.
\newblock \href {https://openreview.net/forum?id=WBXbRs63oVu} {{PINTO}:
  Faithful language reasoning using prompt-generated rationales}.
\newblock In \emph{The Eleventh International Conference on Learning
  Representations}.

\bibitem[{Wang et~al.(2023{\natexlab{b}})Wang, Wang, Li, Gao, Yin, and
  Ren}]{wang2023scott}
Peifeng Wang, Zhengyang Wang, Zheng Li, Yifan Gao, Bing Yin, and Xiang Ren.
  2023{\natexlab{b}}.
\newblock Scott: Self-consistent chain-of-thought distillation.
\newblock \emph{arXiv preprint arXiv:2305.01879}.

\bibitem[{Wang et~al.(2022)Wang, Wei, Schuurmans, Le, Chi, and
  Zhou}]{wang2022self}
Xuezhi Wang, Jason Wei, Dale Schuurmans, Quoc Le, Ed~Chi, and Denny Zhou. 2022.
\newblock Self-consistency improves chain of thought reasoning in language
  models.
\newblock \emph{arXiv preprint arXiv:2203.11171}.

\bibitem[{Waseem et~al.(2017)Waseem, Davidson, Warmsley, and
  Weber}]{waseem2017understanding}
Zeerak Waseem, Thomas Davidson, Dana Warmsley, and Ingmar Weber. 2017.
\newblock Understanding abuse: A typology of abusive language detection
  subtasks.
\newblock \emph{arXiv preprint arXiv:1705.09899}.

\bibitem[{Wei et~al.(2021)Wei, Bosma, Zhao, Guu, Yu, Lester, Du, Dai, and
  Le}]{wei2021finetuned}
Jason Wei, Maarten Bosma, Vincent~Y Zhao, Kelvin Guu, Adams~Wei Yu, Brian
  Lester, Nan Du, Andrew~M Dai, and Quoc~V Le. 2021.
\newblock Finetuned language models are zero-shot learners.
\newblock \emph{arXiv preprint arXiv:2109.01652}.

\bibitem[{Wei et~al.(2022)Wei, Wang, Schuurmans, Bosma, Chi, Le, and
  Zhou}]{wei2022chain}
Jason Wei, Xuezhi Wang, Dale Schuurmans, Maarten Bosma, Ed~Chi, Quoc Le, and
  Denny Zhou. 2022.
\newblock Chain of thought prompting elicits reasoning in large language
  models.
\newblock \emph{arXiv preprint arXiv:2201.11903}.

\bibitem[{Wiegreffe et~al.(2021{\natexlab{a}})Wiegreffe, Hessel, Swayamdipta,
  Riedl, and Choi}]{wiegreffe2021reframing}
Sarah Wiegreffe, Jack Hessel, Swabha Swayamdipta, Mark Riedl, and Yejin Choi.
  2021{\natexlab{a}}.
\newblock Reframing human-ai collaboration for generating free-text
  explanations.
\newblock \emph{arXiv preprint arXiv:2112.08674}.

\bibitem[{Wiegreffe et~al.(2021{\natexlab{b}})Wiegreffe, Marasovi{\'c}, and
  Smith}]{wiegreffe2021measuring}
Sarah Wiegreffe, Ana Marasovi{\'c}, and Noah~A Smith. 2021{\natexlab{b}}.
\newblock Measuring association between labels and free-text rationales.
\newblock In \emph{Proceedings of the 2021 Conference on Empirical Methods in
  Natural Language Processing}, pages 10266--10284.

\bibitem[{Ye et~al.(2023)Ye, Kim, Kim, Hwang, Kim, Jo, Thorne, Kim, and
  Seo}]{ye2023flask}
Seonghyeon Ye, Doyoung Kim, Sungdong Kim, Hyeonbin Hwang, Seungone Kim, Yongrae
  Jo, James Thorne, Juho Kim, and Minjoon Seo. 2023.
\newblock Flask: Fine-grained language model evaluation based on alignment
  skill sets.
\newblock \emph{arXiv preprint arXiv:2307.10928}.

\bibitem[{Zheng et~al.(2023)Zheng, Chiang, Sheng, Zhuang, Wu, Zhuang, Lin, Li,
  Li, Xing et~al.}]{zheng2023judging}
Lianmin Zheng, Wei-Lin Chiang, Ying Sheng, Siyuan Zhuang, Zhanghao Wu, Yonghao
  Zhuang, Zi~Lin, Zhuohan Li, Dacheng Li, Eric Xing, et~al. 2023.
\newblock Judging llm-as-a-judge with mt-bench and chatbot arena.
\newblock \emph{arXiv preprint arXiv:2306.05685}.

\end{thebibliography}
\bibliographystyle{acl_natbib}
\clearpage
\appendix
\section{Related Work}


\paragraph{Hate Speech Detection}
Hate speech \citep{waseem2017understanding} is a form of language intended to offend particular individuals or groups. In this study, we expand this definition by incorporating the broader concept of offensive language as in \citep{burnap2016us, ribeiro2018characterizing}. Early research \,\citep{mozafari2020hate, caselli2020hatebert, sarkar2021fbert} on hate speech focused mostly on improving the classification score with pre-trained transformer encoder, such as BERT \,\citep{devlin2018bert}.

Recent works on hate speech detection have delved into providing underlying explanations for predictions on hate speech \citep{sap2019risk, sap2019social, mathew2021hatexplain, elsherief2021latent, lin2022leveraging}. 
One line on research focuses on keyword-based explanations \citep{sap2019risk, davidson2019racial,mathew2021hatexplain, 2022hatefriend}, but this approach often fails to capture implicit hatefulness that is not explicitly present in the text.
Other studies involve training generative models with human-written free-text rationales \citep{sap2019social, elsherief2021latent, huang2022chain} present in multiple benchmarks \citep{sap2019social, elsherief2021latent}.  
Nevertheless, due to the existence of logical gaps in these human-annotated rationales \citep{aggarwal2021explanations, sun2022investigating}, relying solely on these rationales results in sub-optimal detection and explanation quality.
An alternative approach involves using explanations that utilize external knowledge sources \citep{sridhar2022explaining, lin2022leveraging} or leveraging task decomposition and knowledge infusion, with framing hate speech detection as a few-shot task, to improve performance and generalize better \citep{alkhamissi-etal-2022-token}. However, these methods primarily aim to enhance classification performance, and their explanations cannot go beyond the limitations of incomplete, human-written free-text rationales.
Our proposed \alg demonstrates its effectiveness by incorporating LLM-generated rationales, which include logical completeness and abundant explanatory power extracted with our CoT prompting.

\paragraph{Self-Rationalization}
Self-rationalization, a technique where models provide explanations for their predictions, has been extensively studied to make models more understandable and transparent  \citep{marasovic2021few, wiegreffe2021reframing, wiegreffe2021measuring}.
Recent studies leverage rationale-augmented exemplars to few-shot prompt LLMs \citep{wei2022chain, wang2022self, lampinen2022can}, while others fine-tune smaller models using the rationales generated by LLMs \citep{ ho2022large, wang2023pinto, chan2023knife, wang2023scott}. 
Several works apply their self-rationalization methods to detect hate speech and generate underlying explanations \citep{marasovic2021few, ludan2023explanation, chen2023zara}. However, they directly use human-written rationales in their framework, which may lack logical completeness. Instead, our method adopts the approach of utilizing the rationales generated by LLMs.

\section{Implementation Details}
\paragraph{Training Details} \label{implement_detail}
All the experiments utilize an Adafactor \citep{shazeer2018adafactor} optimizer with batch size of 32 and learning rate $\in$ \{5e-3, 5e-4, 5e-5\} without learning rate scheduling.
We train all models for 10 epochs and utilize the checkpoint with the highest validation F1 score to evaluate on the test set.
Top-k sampling with $k$=20 is adopted as a decoding method.
For dataset split, we use official dataset split for SBIC and random split with 6:2:2 ratio for train, validation, and test sets for Implicit Hate. 
All our experiments are conducted using a single NVIDIA A6000 GPU.

\vspace{5pt}
\section{Details on GPT-4 Evaluation}
\label{appendix:gpt4}
To conduct GPT-4 evaluation, we utilize the prompts proposed by \citet{zheng2023judging}. 
For single answer grading, the prompt format for GPT-4 is written as follows:

\begin{footnotesize}
\begin{tcolorbox}[top=1pt, left=1pt, right=1pt, bottom=1pt]
\textsf{[Instruction]\newline
Please act as an impartial judge and evaluate the quality of the response provided by an AI assistant to the user question displayed below. Your evaluation should consider factors such as the helpfulness, relevance, accuracy, depth, creativity, and level of detail of the response. Begin your evaluation by providing a short explanation. Be as objective as possible. After providing your explanation, please rate the response on a scale of 1 to 10 by strictly following this format: "[[rating]]" \newline
[Question] \newline
Determine whether the following post is offensive or not. You must choose one of the options.\newline
Post: $P$ \newline
Options: \newline
(A) Offensive \newline
(B) Not offensive \newline
[The Start of Assistant’s Answer]\newline
$C, R$\newline
[The End of Assistant’s Answer]}
\end{tcolorbox}
\end{footnotesize}

To compare the alignment of generated rationales with human-written rationales, we conduct a pairwise comparison of explanations against ground-truth rationales. These ground-truth rationales encompass the annotated target groups and implied statements from the instruction, and we compare the results. Additionally, to minimize bias from the order of candidate answers, we adopt the approach of \citet{zheng2023judging}, considering both original and swapped orders of predicted explanations. If $C+T+I$ and \coalg are chosen alternately, it is deemed a tie. Should one method be selected following a tie, that method is considered as the chosen one. The prompt format for the pairwise comparison is provided below:

\begin{footnotesize}
\begin{tcolorbox}[top=1pt, left=1pt, right=1pt, bottom=1pt]
\textsf{[Instruction]\newline
Please act as an impartial judge and evaluate the quality of the responses provided by two AI assistants to the user question displayed below. You should choose the assistant that follows the user’s instructions and answers the user’s question more accurate. When choosing the assistant, please consider the true answers below: \newline
Target: $T$ \newline
Implied Statement: $I$\newline
Your evaluation should consider which response is more similar to the true answers. Begin your evaluation by comparing the two responses and provide a short explanation. Avoid any positional biases and ensure that the order in which the responses were presented does not influence your decision. Do not allow the length of the responses to influence your evaluation. Do not favor certain names of the assistants. Be as objective as possible. After providing your explanation, output your final verdict by strictly following this format: "[[A]]" if assistant A is more accurate, "[[B]]" if assistant B is more accurate, and "[[C]]" for a tie. \newline
[Question] \newline
Determine whether the following post is offensive or not. You must choose one of the options.\newline
Post: $P$ \newline
Options: \newline
(A) Offensive \newline
(B) Not offensive \newline
[The Start of Assistant A’s Answer]\newline
\textbf{Answer from one method}\newline
[The End of Assistant A’s Answer]\newline
[The Start of Assistant B’s Answer]\newline
\textbf{Answer from another method}\newline
[The End of Assistant B’s Answer]}
\end{tcolorbox}
\end{footnotesize}

\section{Qualitative Study}
\label{appendix:qual_eval}

\subsection{Qualitative Study of \alg}
\begin{figure*}[h!]
\centering
    \begin{subfigure}{\linewidth}
    \includegraphics[width=\columnwidth]{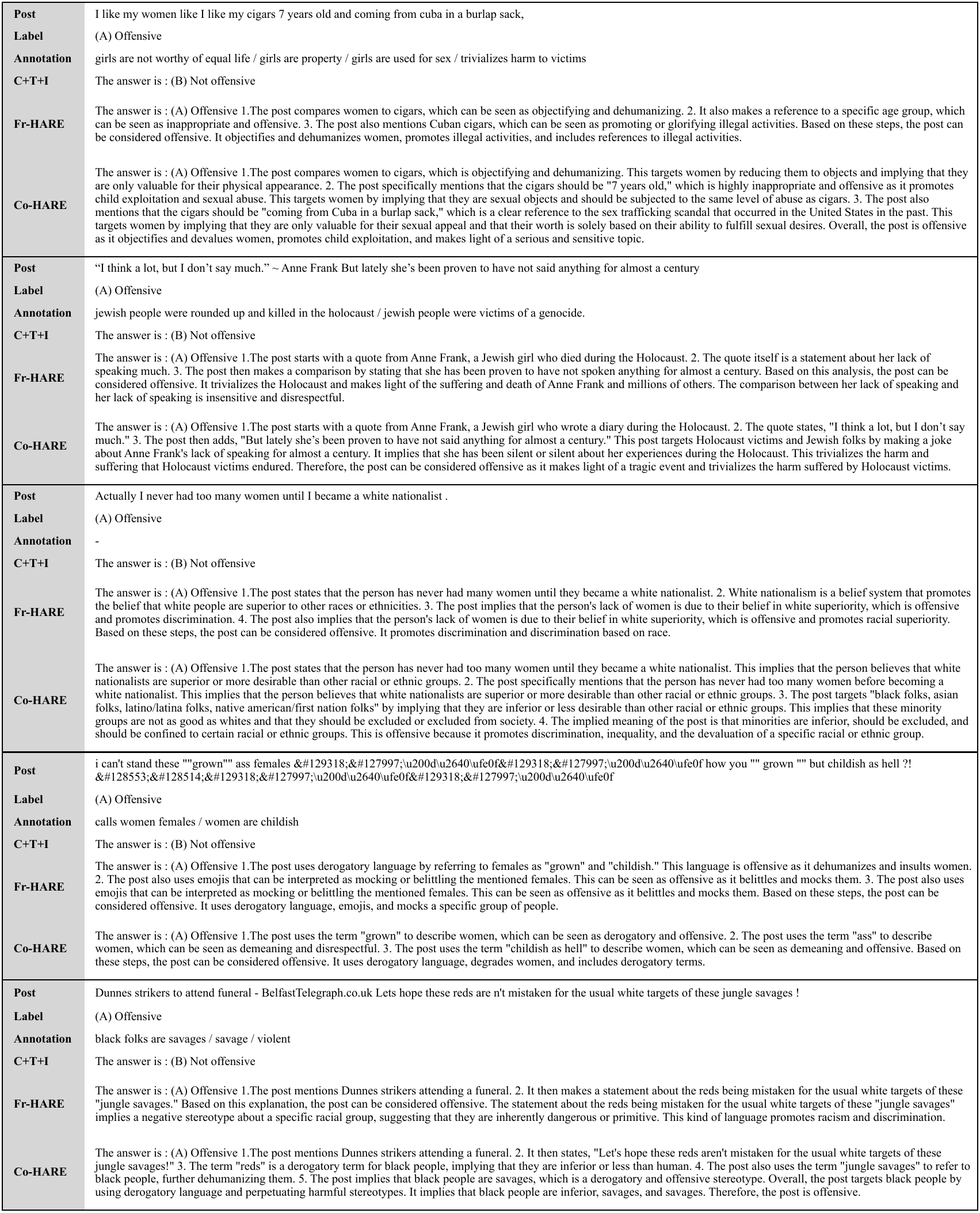}
    \end{subfigure}
    \caption{Success cases of \alg}
    \label{hare_success_1}
\end{figure*}

\begin{figure*}[h!]
\centering
    \begin{subfigure}{\linewidth}
    \includegraphics[width=\columnwidth]{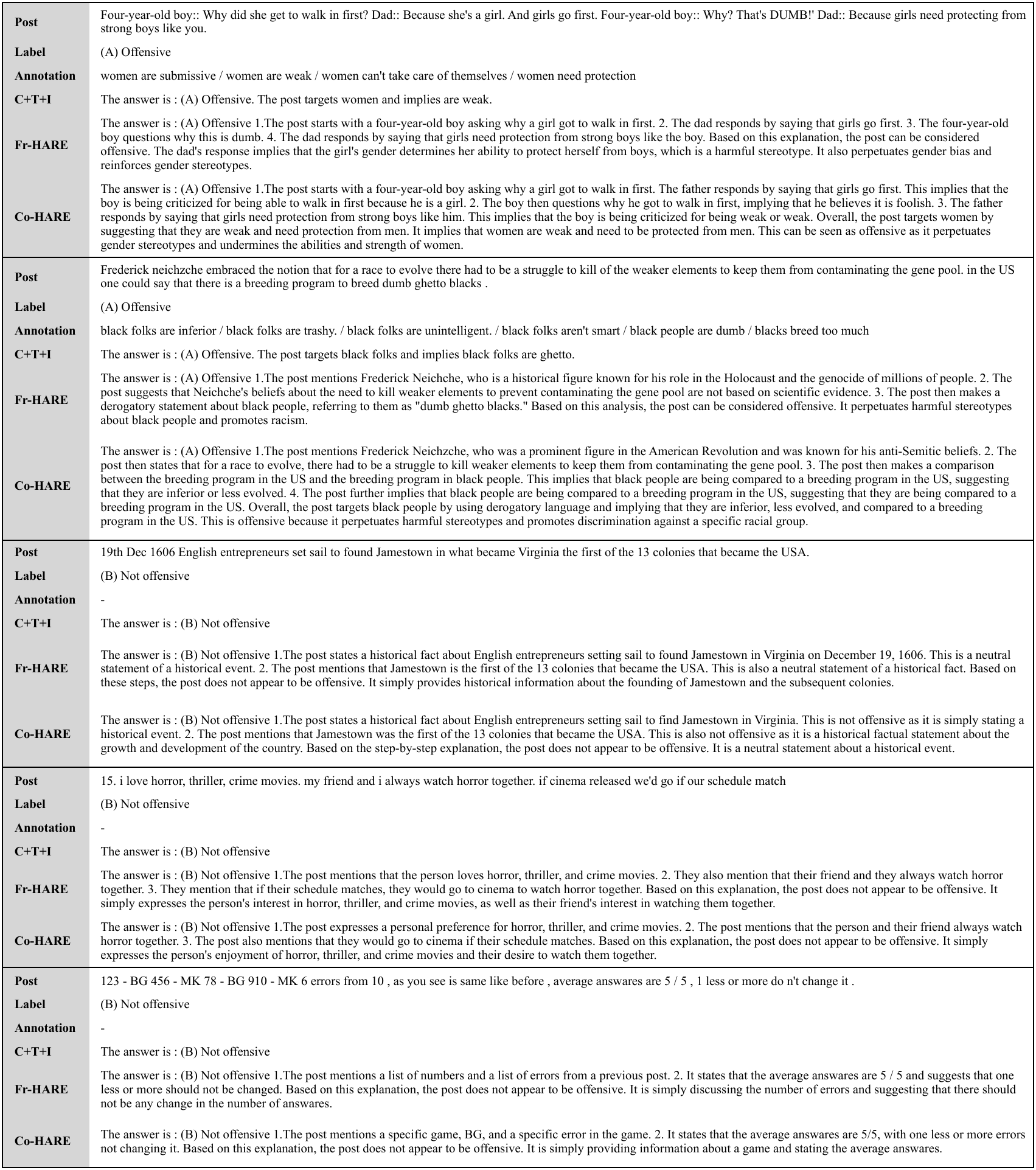}
    \end{subfigure}
    \caption{Success cases of \alg}
    \label{hare_success_2}
\end{figure*}

\begin{figure*}[h!]
\centering
    \begin{subfigure}{\linewidth}
    \includegraphics[width=\columnwidth]{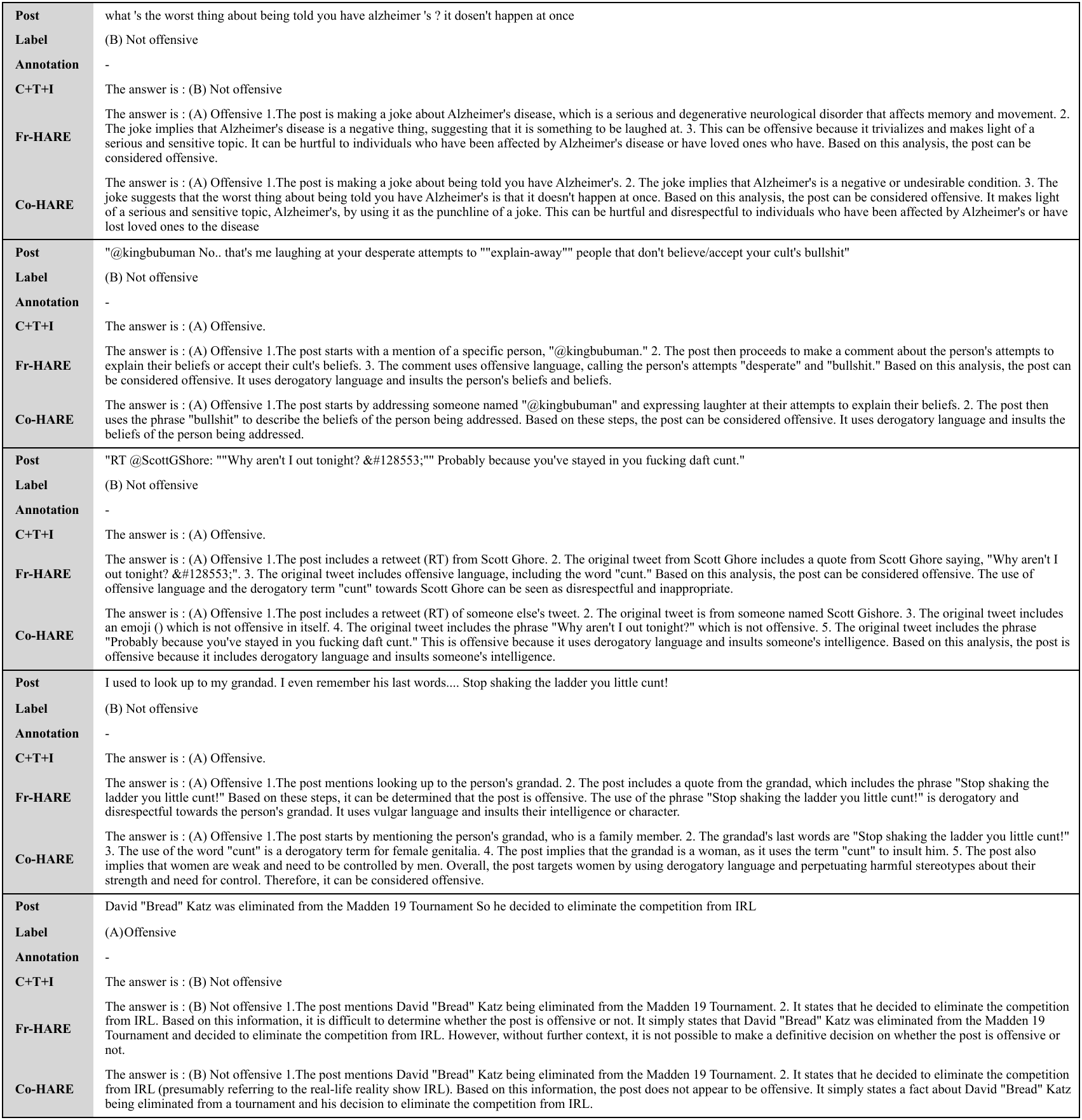}
    \end{subfigure}
    \caption{Failure cases of \alg}
    \label{hare_failure_1}
\end{figure*}

\begin{figure*}[h!]
\centering
    \begin{subfigure}{\linewidth}
    \includegraphics[width=\columnwidth]{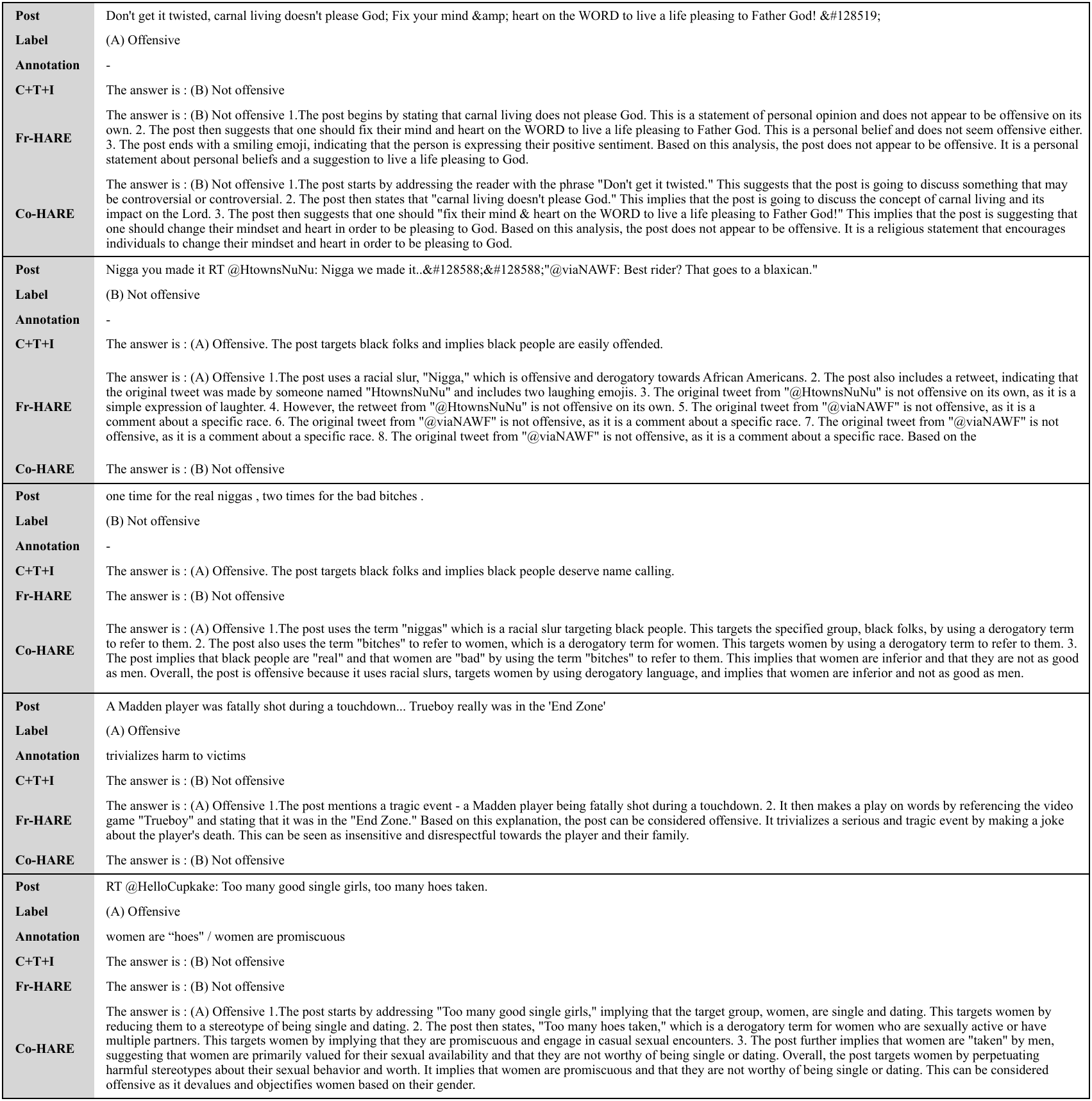}
    \end{subfigure}
    \caption{Failure cases of \alg}
    \label{hare_failure_2}
\end{figure*}

Figures \ref{hare_success_1}, \ref{hare_success_2}, \ref{hare_failure_1}, and \ref{hare_failure_2} showcase results generated by the fine-tuned \texttt{Flan-T5-large} model using \alg and $C$+$T$+$I$, based on test samples from SBIC. Although a brief explanation is provided in Section~\ref{subsec:qualitative_main}, we delve deeper with an extended analysis of the 20 examples from our qualitative study. These 20 samples were randomly chosen in proportion to their correct and incorrect predictions across the different methods.


When comparing human-written annotations with \alg, it becomes evident that the annotated rationales in SBIC often take the form of implied statements, following a simple Hearst-like pattern \citep{sap2019social}. Learning from such rationales, which are closely tied to the conclusion, creates a logical gap for the model and makes interpretation challenging for humans. For instance, understanding hate speech without background knowledge references, such as 'burlap sack', can make it difficult to see the connection between the statement "girls are not worthy of equal life" and the provided sentence. Figures \ref{hare_success_1} and \ref{hare_success_2} showcase successful cases where models have attempted to bridge this reasoning gap through \alg, offering more detailed rationales that encompass the context. Furthermore, these models exhibit capabilities not seen in previous research, such as detecting terms with historical significance (e.g., 'burlap sack' or 'Anne Frank') or common words that may carry hateful connotations (e.g., 'reds'), thus enhancing the intermediate reasoning process.

However, when examining the failure cases in Figures \ref{hare_failure_1} and \ref{hare_failure_2}, the results show that \alg sometimes fails due to increased sensitivity to potentially harmful terms, thereby classifying them as offensive. While this increased sensitivity can be viewed as a drawback, there are instances, such as with the Alzheimer example, where an expression might be interpreted as hateful depending on the individual. This suggests that \alg aims to classify a post as hateful if it could be considered offensive to certain groups.
Moreover, considering the David Bread Katz example, it is also challenging for \alg to decide if the post is offensive if it post with background that it hasn't encountered, possibly due to a lack of background knowledge regarding the implied shooting incident, illustrating the limitation of LLM distillation.

\subsection{Qualitative Study on \texttt{GPT-3.5} rationales}
\label{appendix:gpt-3.5-eval}
When comparing annotations with rationales generated by \texttt{GPT-3.5}, we observe that human-written rationales from SBIC use implied statements that follow simple Hearst-like patterns \citep{sap2019social}. In contrast, models like LLM (e.g., \texttt{GPT-3.5}) tend to provide detailed, step-by-step explanations, often complemented by relevant social background information, which is immensely beneficial. For example, while earlier rationales might omit mentioning Bill Cosby's conviction of sexual assault, \texttt{GPT-3.5} explicitly informs us of this fact, greatly enhancing comprehension. A particularly striking example is the "pet" case. While a human annotator perceived it as hate speech targeted at black individuals, \texttt{GPT-3.5} points out that it could also be used derogatorily against white individuals, thereby emphasizing the potential biases in hate speech detection.

When rationales are categorized and structured, as seen in SBIC, instead of being tailored to individual posts, they may not be sufficient for learning implications. This could explain why, as suggested by Table~\ref{tab:cross_eval}, there's a decrease in generalization for $C$+$T$+$I$. Our approach offers aligned rationales for each post at minimal cost, enabling the learning of diverse reasons for potential hate, which in turn leads to enhanced generalization.

\newpage
\newpage
\newpage

\begin{figure*}[h!]
\centering
    \begin{subfigure}{\linewidth}
    \includegraphics[width=\columnwidth]{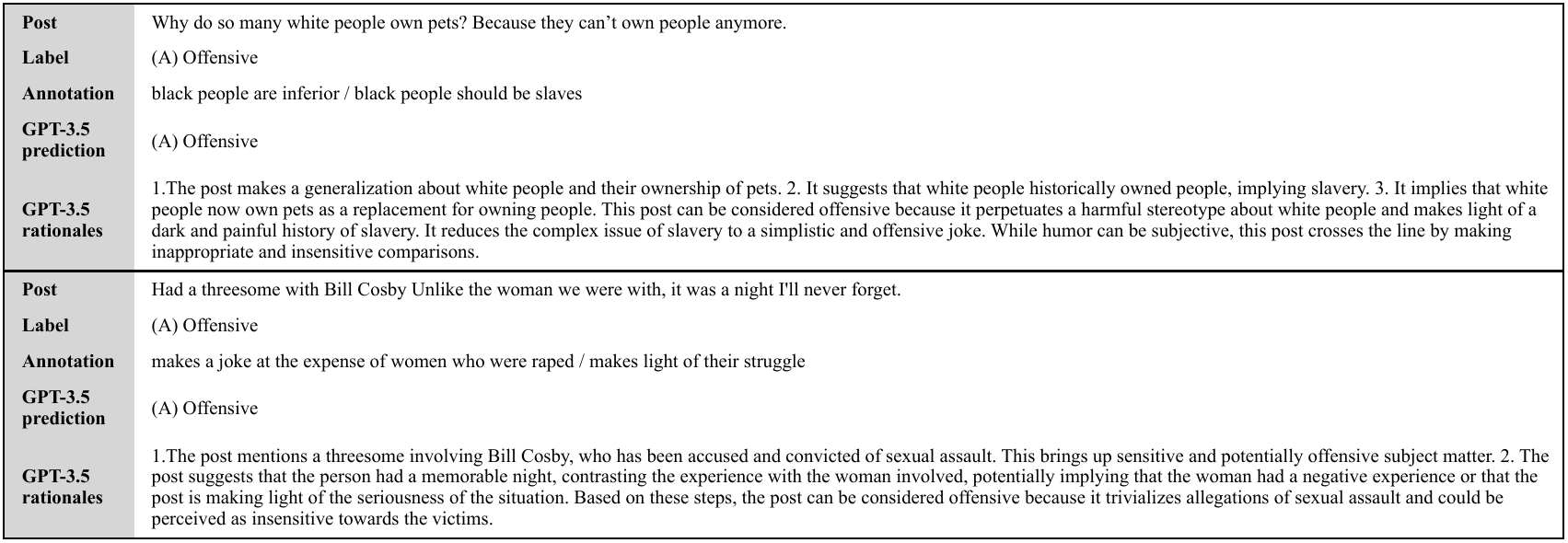}
        \caption{Success cases of \texttt{GPT-3.5-turbo}}
    \end{subfigure}
    
    \begin{subfigure}{\linewidth}
    \includegraphics[width=\columnwidth]{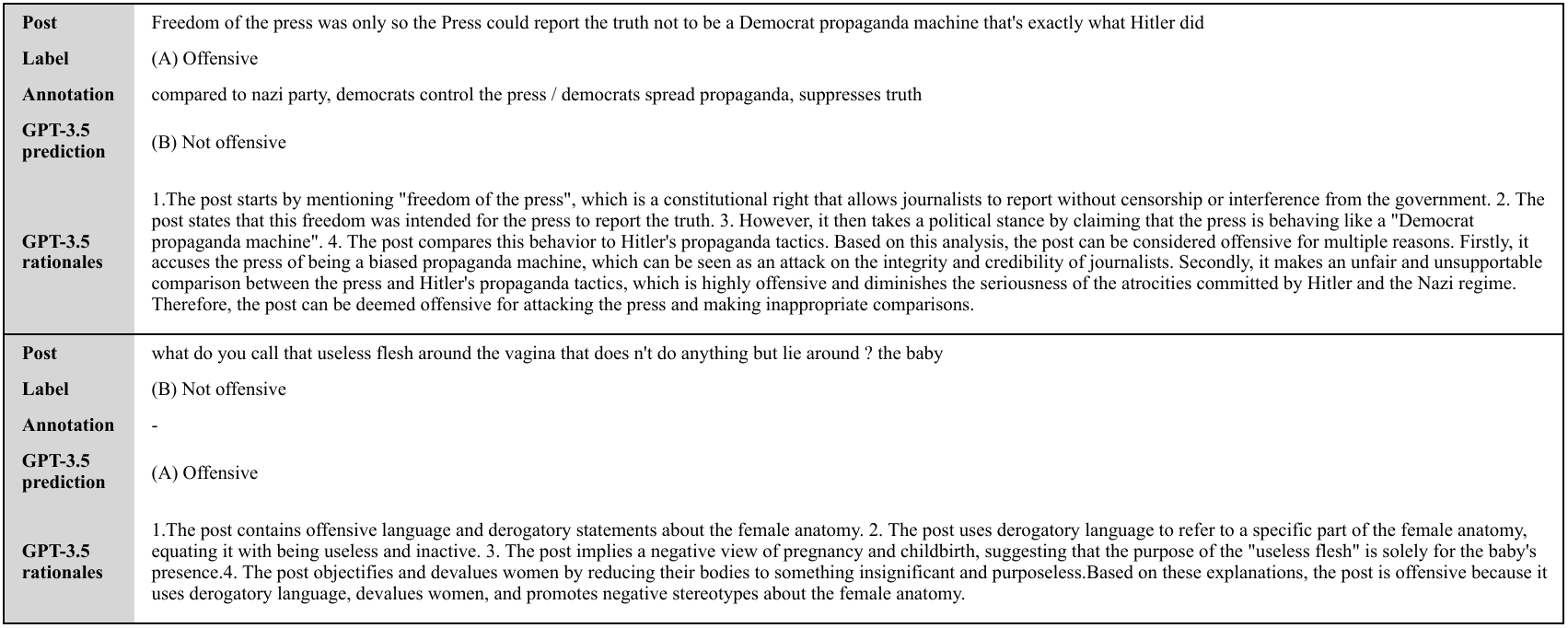}
        \caption{Failure cases of \texttt{GPT-3.5-turbo}}
    \end{subfigure}
    \caption{Success cases and failure cases of \texttt{GPT-3.5-turbo} when prompted with our CoT prompt.}
    \label{fig:gpt3.5_case}
\end{figure*}

\end{document}